\documentclass[runningheads]{llncs}
\usepackage{graphicx}
\usepackage{amsmath,amssymb} 
\usepackage{color}
\usepackage{multirow}
\usepackage{epsfig}
\usepackage{graphicx}
\usepackage{booktabs}
\usepackage{colortbl}
\usepackage{xcolor}
\usepackage{graphicx}
\usepackage{lipsum}
\usepackage{hyperref}

\definecolor{tablecolor}{rgb}{0.0,0.0,0.0}
\definecolor{cwblue1}{rgb}{0.27,0.427,0.623}
\definecolor{cwblue2}{rgb}{0.286,0.454,0.658}
\definecolor{cwblue3}{rgb}{0.733,0.811,0.905}

\newcommand{\myparagraph}[1]{\textbf{#1}}  

\newcommand{\mB}{\mbox{$\mathbf{B}$}}

\newcommand{\mO}{\mbox{$\mathbf{O}$}}

\newcommand{\mR}{\mbox{$\mathbf{R}$}}

\newcommand{\mU}{\mbox{$\mathbf{U}$}}
\newcommand{\mV}{\mbox{$\mathbf{V}$}}
\newcommand{\mW}{\mbox{$\mathbf{W}$}}
\newcommand{\mX}{\mbox{$\mathbf{X}$}}

\newcommand{\mZ}{\mbox{$\mathbf{Z}$}}

\newcommand{\mb}{\mbox{$\mathbf{b}$}}
\newcommand{\mc}{\mbox{$\mathbf{c}$}}

\newcommand{\mg}{\mbox{$\mathbf{g}$}}
\newcommand{\mh}{\mbox{$\mathbf{h}$}}

\newcommand{\mo}{\mbox{$\mathbf{o}$}}

\newcommand{\mr}{\mbox{$\mathbf{r}$}}
\newcommand{\ms}{\mbox{$\mathbf{s}$}}

\newcommand{\mathu}{\mbox{$\mathbf{u}$}}
\newcommand{\mv}{\mbox{$\mathbf{v}$}}

\newcommand{\my}{\mbox{$\mathbf{y}$}}

\newcommand{\etal}{\textit{et al.}}

\newcommand{\mycomment}[1]{}

\begin{document}
\pagestyle{headings}
\mainmatter
\def\ECCV18SubNumber{}  

\title{Object Level Visual Reasoning in Videos} 

\titlerunning{Object Level Visual Reasoning in Videos}

\authorrunning{Baradel \etal}

\author{Fabien Baradel\inst{1}
\and Natalia Neverova\inst{2}
\and Christian Wolf\inst{1,3}\and \\Julien Mille\inst{4} \and Greg Mori\inst{5}}

\institute{ Universit\'e Lyon,  INSA Lyon, CNRS, LIRIS, F-69621, Villeurbanne, France, \email{firstname.lastname@liris.cnrs.fr} \and
			Facebook AI Research, Paris, France, \email{nneverova@fb.com} \and
		    INRIA, CITI Laboratory, Villeurbanne, France \and
		    Laboratoire d'Informatique de l'Univ. de Tours, INSA Centre Val de Loire, \\41034, Blois, France, \email{julien.mille@insa-cvl.fr} \and
			Simon Fraser University, Vancouver, Canada, \email{mori@cs.sfu.ca}
			\\~\\
			\url{https://fabienbaradel.github.io/eccv18_object_level_visual_reasoning/}
			}


\maketitle

\begin{abstract}
Human activity recognition is typically addressed by 
detecting key concepts like global and local motion, features related
to object classes present in the scene, as well as features related to
the global context.  The next open challenges in activity recognition
require a level of understanding that pushes beyond this and call for models with capabilities for fine distinction and detailed comprehension of interactions
between actors and objects in a scene.  We propose a model capable of
learning to reason about semantically meaningful spatio-temporal
interactions in videos. The key to our approach is a choice of
performing this reasoning at the object level through the integration
of state of the art object detection networks. This allows
the model to learn detailed spatial interactions that exist at a
semantic, object-interaction relevant level.
We evaluate our method on three standard datasets (Twenty-BN Something-Something, VLOG and EPIC Kitchens)  and achieve state of the art results on all of them.
Finally, we show visualizations of the interactions learned by the 
model, which illustrate object classes and their interactions 
corresponding to different activity classes.
\end{abstract}

\section{Introduction}
\label{sec:introduction}



The field of video understanding is extremely diverse, ranging from extracting highly detailed information captured by specifically designed motion capture systems~\cite{Shahroudy2016} to making general sense of videos from the Web~\cite{Youtube8M2016}.
As in the domain of image recognition, there exist a number of large-scale video datasets~\cite{Carreira_2017_CVPR,Monfort_2018_arxiv_moments,Goyal_2017_ICCV,Fouhey_2017_arxiv_b,KrishnaVG17,46331}, which allow the training of high-capacity deep learning models from massive amounts of data.  These models enable detection of key cues present in videos, such as global and local motion, various object categories and global scene-level information, and often achieve impressive performance in recognizing high-level, abstract concepts in the wild.  

\mycomment{
Video understanding is a diverse field in computer vision.  At one end of the spectrum is video classification, with a goal of recognition of high level concepts in diverse internet video, exemplified by the YouTube-8M dataset~\cite{Youtube8M2016}.  On the other is close range 3D human activity recognition based on specialized RGB-D sensors and laboratory data collection~\cite{Shahroudy2016}.   Powerful models exist for analyzing internet videos.  As in object detection, large-scale datasets have been introduced \cite{Carreira_2017_CVPR,Monfort_2018_arxiv_moments,Goyal_2017_ICCV,Fouhey_2017_arxiv_b,KrishnaVG17,46331}, which allow the training of high-capacity models from large amounts of data.  These models enable the detection of key concepts present in videos, capturing concepts such as global and local motion, the presence of various object categories in a video, and global scene-level information.  Impressive performance in recognizing high-level, often abstract concepts in diverse internet videos has been achieved.
}

However, recent attention has been directed toward a more thorough understanding of human-focused activity in diverse internet videos.  These efforts range from atomic human actions~\cite{46331} to fine-grained object interactions~\cite{Goyal_2017_ICCV} to everyday, commonly occurring human-object interactions~\cite{Fouhey_2017_arxiv_b}.  This returns us to a human-centric viewpoint of activity recognition where it is not only the presence of certain objects / scenes that dictate the activity present, but the manner, order, and effects of human interaction with these scene elements that are necessary for understanding.  In a sense, this is akin to the problems in current 3D human activity recognition datasets~\cite{Shahroudy2016}, but requires the more challenging reasoning and understanding of diverse environments common to internet video collections.



\begin{figure}[t]
	\centering
	\includegraphics[width=11cm]{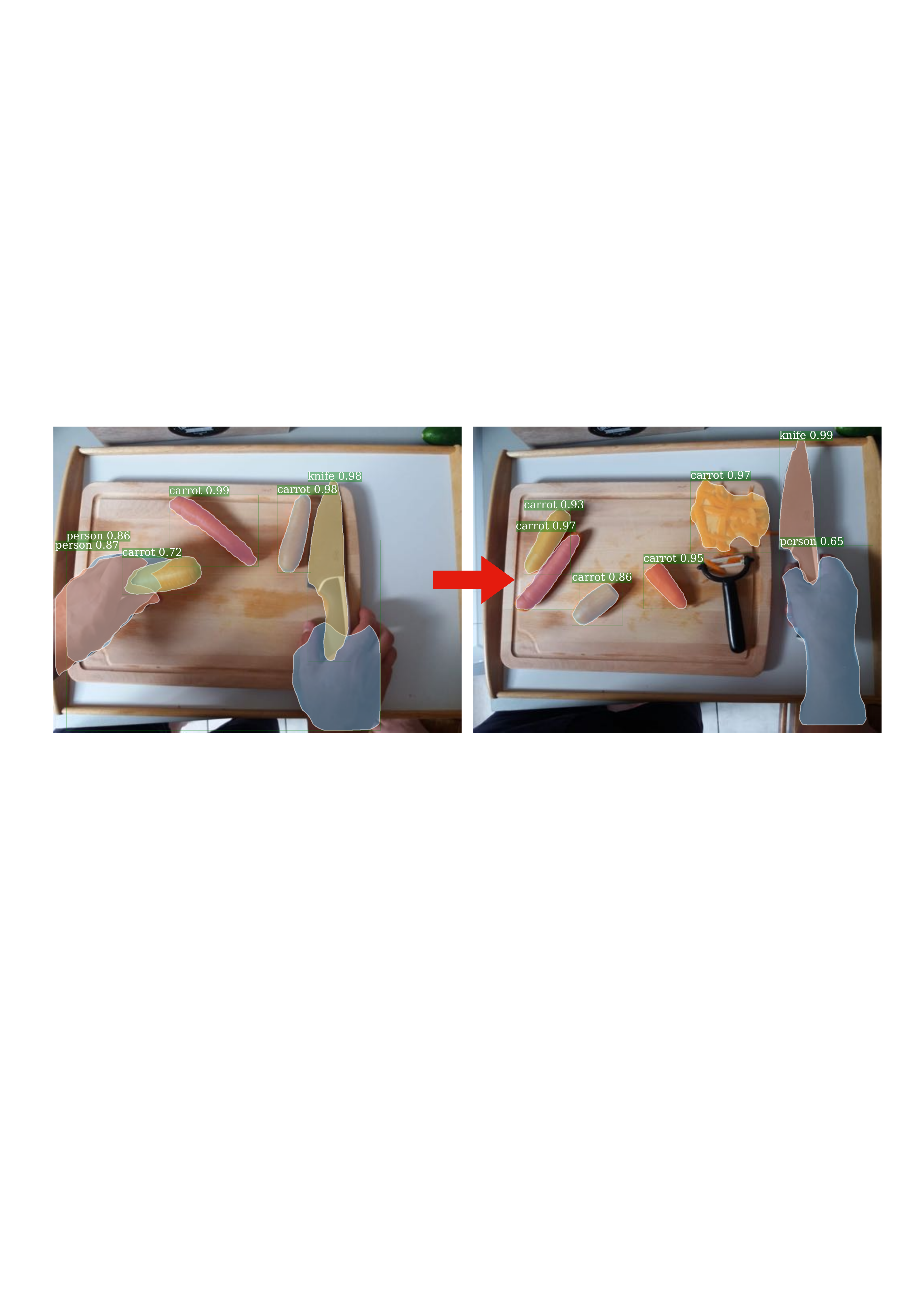} \\
	\caption{Humans can understand what happened in a video (``the leftmost carrot was chopped by the person'') given only a pair of frames. Along these lines, the goal of this work is to explore the capabilities of higher-level \textit{reasoning} in neural models operating at the semantic level of objects and interactions.}
	\label{fig:rolling_car}
\end{figure}

Humans are able to infer what happened in a video given only a few sample frames.
In particular, they can infer complex activities happening between pairs of frames.
This faculty is called \textit{reasoning} and is a key component of human intelligence.
As an example we can consider the pair of images in 
Figure~\ref{fig:rolling_car}, which shows a complex situation involving articulated objects (human, carrots and knife), the change of location and composition of objects.
For humans it is straightforward to draw a conclusion on what happened (a carrot was chopped by the human).
Humans have this extraordinary ability of performing visual reasoning on very complicated tasks while it remains unattainable for contemporary computer vision algorithms~\cite{Stabinger-2016-ICANN,Fleuret2011ComparingMA}.


The ability to perform \textit{visual reasoning} in computer vision algorithms is still an open problem. Attempts have been made for learning interactions between different entities in images with promising results on Visual Question Answering.
There have been a number of attempts to equip neural models with reasoning abilities by training them to solve Visual Question Answering (VQA) problems.
Among proposed solutions are prior-less data normalization~\cite{PerezCourville_ICML2017WS}, structuring networks to model relationships~\cite{Santoro_2107_NIPS,Watters_2017_NIPS} as well as more complex attention based mechanisms \cite{HudsonManningICLR2018MacCells}. At the same time, it was shown that high performance on existing VQA datasets can be achieved by simply discovering biases in the data~\cite{kim2018notsoclevr}.



We extend these efforts to \textit{object level reasoning in videos}.
Since a video is a temporal sequence, we leverage time as an explicit causal signal to identify causal object relations.
Our approach is related to the concept of the \textit{``arrow of the time"}~\cite{Pickup_2014_CVPR} involving the ``one-way direction'' or ``asymmetry'' of time.
Causal event occurs before the event it affects ($A \rightarrow B$).
In Figure~\ref{fig:rolling_car} the knife was used before the carrot switched over to the chopped-up state on the right side.
For a video classification problem, we want to identify a causal event $A$ happening in a video that affects its label $B$.
But instead of identifying this causal event directly from pixels we want to identify it from an object level perspective.
We believe that such an approach would be able to learn causal signals.

%

Following this hypothesis we propose to make a bridge between object detection and activity recognition.
Object detection allows us to extract low-level information from a scene with all the present object instances and their semantic meanings.
However, detailed activity understanding requires reasoning over these semantic structures, determining which objects were involved in interactions, of what nature, and what were the results of these.  To compound problems, the
semantic structure of a scene may change during a video (e.g. a new object can appear, a person may make a move from one point to another one of the scene).


We propose an \textbf{Object Relation Network} (ORN), a neural network module for reasoning between detected semantic object instances through space and time.  The ORN has potential to address these issues and conduct relational reasoning over object interactions for the purpose of activity recognition.  A set of object detection masks ranging over different object categories and temporal occurrences is input to the ORN.
The ORN is able to infer pairwise relationships between objects detected at varying different moments in time.

Code and object masks predictions will be publicly available\footnote{\url{https://github.com/fabienbaradel/object_level_visual_reasoning}}.




\section{Related work}
\label{sec:related}


\myparagraph{Action Recognition.}
Action recognition has a long history in computer vision.
Pre-deep learning approaches in action recognition focused on handcrafted spatio-temporal features including space-time interest points like SIFT-3D, HOG3D, IDT and aggregated them using bag-of-words techniques. Some hand-crafted representations, like dense trajectories~\cite{WangSchmid2011}, still give competitive performance and are frequently combined with deep learning.

In the recent past, work has shifted to deep learning. Early attempts adapt 2D convolutional networks to videos through temporal pooling and 3D convolutions~\cite{Baccouche2011,Tran_2015_ICCV}.
3D convolutions are now widely adopted for activity recognition with the introduction of feature transfer by inflating pre-trained 2D convolutional kernels from image classification models trained on ImageNet/ILSVRC~\cite{Russakovsky2015} through 3D kernels~\cite{Carreira_2017_CVPR}.
The downside of 3D kernels is their computational complexity and the large number of learnable parameters, leading to the introduction of 2.5D kernels, i.e. separable filters in the form of a 2D spatial kernel followed by a temporal kernel~\cite{Xie_arxiv_2017_a}. An alternative to temporal convolutions are Recurrent Neural Networks (RNNs) in their various gated forms (GRUs, LSTMs)~\cite{Hochreiter1997,lrcn2014}.

Karpathy \etal~\cite{KarpathyCVPR14} presented a wide study on different ways of connecting information in spatial and temporal dimensions through convolutions and pooling. On very general datasets with coarse activity classes they have showed that there was a small margin between classifying individual frames and classifying videos with more sophisticated temporal aggregation.

Simoyan \etal~\cite{NIPS2014_5353} proposed a widely adopted two-stream architecture for action recognition which extracts two different streams, one processing raw RGB input and one processing pre-computed optical flow images. 
The method outperformed the state of the art, but relies on rather small scale optical flow computations.

In slightly narrower settings, prior information on the video content can allow more fine-grained models. Articulated pose is widely used in cases where humans are guaranteed to be present~\cite{Shahroudy2016}. Pose estimation and activity recognition as a joint (multi-task) problem has recently shown to improve both tasks \cite{Luvizon_2018_CVPR}. 
Somewhat related to our work, Structural RNNs \cite{JainStructuralRNN2016} perform activity recognition by integrating features from semantic objects and their relationships. 
However, they handle the temporal evolution of tracked objects in videos with a set of RNNs, each of which corresponds to cliques in a graph which models the spatio-temporal relationships between these objects. This graph is hand-crafted manually for each application, though related work provides learnable connections via gating functions~\cite{DengVHM16}.

Attention models are a way to structure deep networks in an often generic way. They are able to iteratively focus attention to specific parts in the data without requiring prior knowledge about part or object positions. In activity recognition, they have gained some traction in recent years, either as soft-attention on articulated pose (joints) \cite{Song2016}, on feature map cells \cite{Sharma2016a,Sun_2017_ICCV}, on time \cite{yeung2015every} or on parts in raw RGB input through differentiable crops \cite{BaradelCVPR2018}.

When raw video data is globally fed into deep neural networks, they focus on extracting spatio-temporal features and perform aggregations. It has been shown that these techniques fail on challenging fine-grained datasets, which require learning long temporal dependencies and human-object interactions. A concentrated effort has been made to create large scale datasets to overcome these issues~\cite{Goyal_2017_ICCV,Fouhey_2017_arxiv_b,KrishnaVG17,46331}.

\myparagraph{Relational Reasoning.}
Relational reasoning is a well studied field for many applications ranging from visual reasoning ~\cite{Santoro_2107_NIPS} to reasoning about physical systems~\cite{Battaglia_2016_NIPS}.
Battaglia \etal~\cite{Battaglia_2016_NIPS} introduce a fully-differentiable network physics engine called Interaction Network (IN).
IN learns to predict several physical systems such as gravitational systems, rigid body dynamics, and mass-spring systems.
It shows impressive results; however, it learns from a virtual environment, which provides access to virtually unlimited training examples.
Following the same perspective, Santoro \etal~\cite{Santoro_2107_NIPS} introduced Relation Network (RN), a plug-in module for  reasoning in deep networks. RN shows human-level performance in Visual Question Answering (VQA) by inferring pairwise ``object'' relations. However, in contrast to our work, the term ``object'' in~\cite{Santoro_2107_NIPS}  does not refer to semantically meaningful entities, but to discrete cells in feature maps. The number of interactions therefore grows with feature map resolutions, which makes it difficult to scale. Furthermore, a recent study~\cite{kim2018notsoclevr} has shown that some of these results are subject to dataset bias and do not generalize well to small changes in the settings of the dataset.

In the same line, a recent work~\cite{vanSteenkiste_2018_ICLR} has shown promising results on discovering objects and their interactions in an unsupervised manner using training examples from virtual environments. In~\cite{velickovic2018graph}, attention and relational modules are combined on a graph structure. From a different perspective, \cite{PerezCourville_ICML2017WS} show that relational reasoning can be learned for visual reasoning in a data driven way without any prior using conditional batch normalization with a feature-wise affine transformation based on conditioning information. In an opposite approach, a strong structural prior is learned in the form of a complex attention mechanism: in~\cite{HudsonManningICLR2018MacCells}, an external memory module combined with attention processes over input images and text questions, performing iterative reasoning for VQA.

While most of the discussed work has been designed for VQA and for predictions on physical systems and environments, extensions have been proposed for video understanding. 
Reasoning in videos on a mask or segmentation level has been attempted for video prediction \cite{LucNeverovaICCV2017}, where the goal was to leverage semantic information to be able predict further into the future.
Zhou et al~\cite{Zhou_2017} have recently shown state-of-the-art performance on challenging datasets by extending Relation Network to video classification. Their chosen entities are frames, on which they employ RN to reason on a temporal level only through  pairwise frame relations. The approach is promising, but restricted to temporal contextual information without an understanding on a local object level, which is provided by our approach.

Reasoning over sets of objects is somewhat related to reasoning from unstructured data points, as done in  
PointNet~\cite{PointNetCVPR2017}, designed to learn from unordered sets of points.
PointNet shares many properties with DeepSet~\cite{NIPS2017_6931} which is a more general framework for extracting information from sets of objects. To some extent, our work is related to PointNet, as we handle unordered sets of objects in a permutation invariant way.  However, we have an object relation viewpoint that directly reasons over relationships between these semantic entities.





\section{Object-level Visual Reasoning in Space and Time}
\label{sec:method}



\begin{figure}[t]
	\centering
	\includegraphics[width=\linewidth]{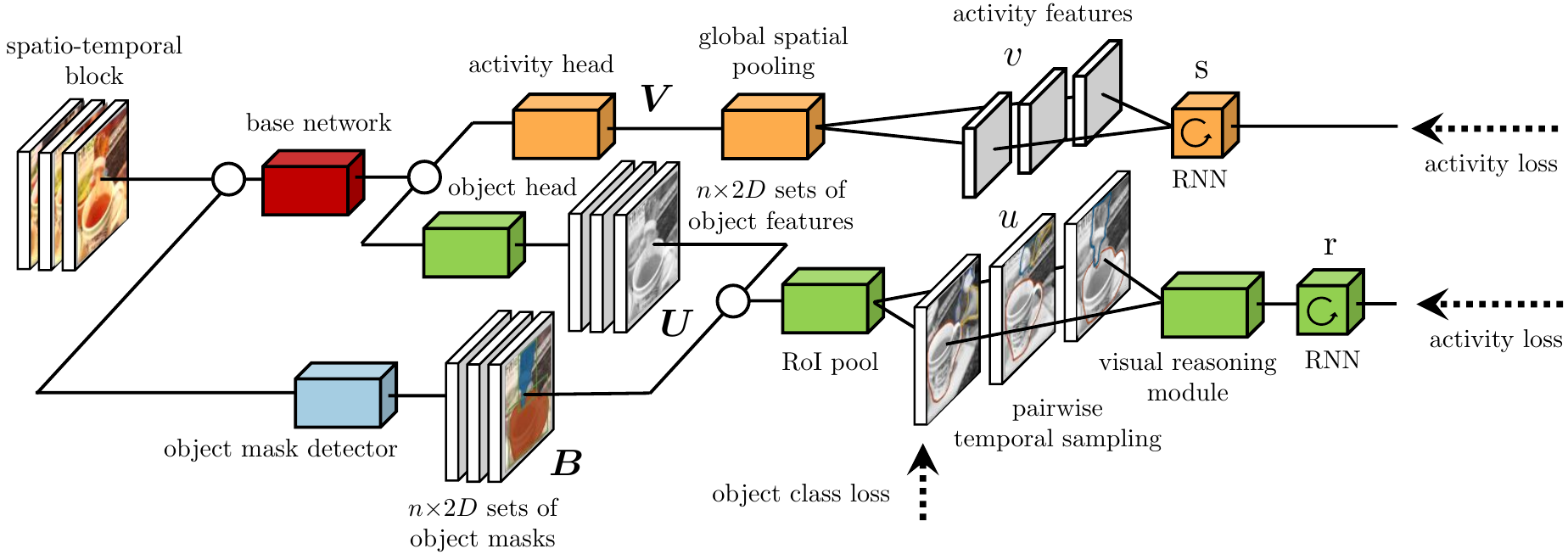}
	\caption{A functional overview of the model. A global convolutional model extracts features and splits into two heads trained to predict, respectively activity classes and object classes. The latter are predicted by pooling over object instance masks, which are predicted by an additional convolutional model. The object instances are passed through a visual reasoning module.}
	\label{fig:block_diagram}
\end{figure}

Our goal is to extract multiple types of cues from a video sequence: interactions between predicted objects and their semantic classes, as well as  local and global motion in the scene. We formulate this objective as a neural architecture with two heads: an \textit{activity head} and an \textit{object head}. 
Figure \ref{fig:block_diagram} gives a functional overview of the model. Both heads share common features up to a certain layer shown in red in the figure.
The \textit{activity head}, shown in orange in the figure, is a CNN-based architecture employing convolutional layers, including spatio-temporal convolutions, able to extract global motion features. However, it is not able to extract information from an object level perspective. We leverage the \textit{object head} to perform reasoning on the relationships between predicted object instances.

Our main contribution is a new structured module called \textbf{Object Relation Network} (ORN), which is able to perform spatio-temporal reasoning between detected object instances in the video.
ORN is able to reason by modeling how objects move, appear and disappear and how they interact between two frames.


In this section, we will first describe our main contribution, the ORN network. We then provide details about object instance features, about the activity head, and finally about the final recognition task.
In what follows, lowercase letters denote 1D vectors while uppercase letters are used for 2D and 3D matrices or higher order tensors. We assume that the input of our system is a video of $T$ frames denoted by $\mX_{1:T} = (\mX_t)_{t=1}^T$ where $\mX_t$ is the RGB image at timestep $t$. The goal is to learn a mapping from $\mX_{1:T}$ to activity classes $\my$.

%

\subsection{Object Relation Network}
\label{sec:sorn}
\noindent
ORN (Object Relation Network) is a module for reasoning between semantic objects through space and time. It captures object moves, arrivals and interactions in an efficient manner. We suppose that for each frame $t$, we have a set of objects $k$ with associated features $\mo_t^k$. Objects and features are detected and computed by the object head described in Section \ref{sec:mask_predictor}. 

Reasoning about activities in videos is inherently temporal, as activities follow the \emph{arrow of time}~\cite{Pickup_2014_CVPR}, i.e. the causality of the time dimension imposes that past actions have consequences in the future but \emph{not} vice-versa. We handle this by sampling: running a process over time $t$, and for each instant $t$, sampling a second frame $t'$ with $t'{<}t$. Our network reasons on objects which interact between pairs of frames and their corresponding sets of objects $\mO_{t'} = \big\{ \mo^k_{t'} \big\}^{K'}_{k=1}$ and $\mO_{t} = \big\{ \mo^k_{t} \big\}^{K}_{k=1}$. The goal is to learn a general function defined on the set of all input objects from the combined set of both frames:
\begin{gather}
\mg_t = g(\mo^1_{t'},\dots,\mo^{K'}_{t'},\mo^1_{t},\dots,\mo^K_{t}).
\end{gather}
The objects in this set are unordered, aside for the frame they belong to.

This task is related to a problem raised in the PointNet algorithm \cite{PointNetCVPR2017} discussed in Section~\ref{sec:related}. PointNet approximates a general function $g$ over an item set ${\cal{S}} = \{x_1,x_2,\dots,x_N\}$ as a symmetric function $g'$ on transformed elements of the set
\begin{gather}
g(x_1,x_2,\dots,x_N) \approx g'(h(x_1),h(x_2),\dots, h(x_N)).
\label{eq:point_net}
\end{gather}
\noindent In \cite{PointNetCVPR2017}, $g'$ is a max pooling operation. The authors show that it allows universal approximation of continuous sets of functions given that the hidden representation (the output of the mapping $h(\cdot)$) is of sufficiently high dimension.

We argue that the approximation in (\ref{eq:point_net}) can be extended as follows:
\begin{gather}
g(x_1,x_2,\dots,x_N) \approx f_\phi
\left (
\sum_{c\in\cal{C}} h_c \left ( \displaystyle{\cup_{i\in c} \ x_i}  \right )
\right )
\label{eq:general_form}
\end{gather}
where $\cal{C}$ is the set of cliques of a graph defined over the item set $\cal{S}$, $\cup$ is the concatenation operator and we chose the sum operator as symmetric function. The input dimension of the non-linearity $h_c(\cdot)$ depends on the size of clique $c$ but maps to a fixed output dimension $H$. In the case where $\cal{C}$ is composed of unary cliques only, form (\ref{eq:general_form}) decomposes like (\ref{eq:point_net}) with the exception of a different symmetry operator (sum instead of max pooling). Choosing different graphical structures through $\cal{S}$ will lead to different terms in the summation and allows modeling different types of interactions between items in the item set.

Note that the interactions between items in (\ref{eq:general_form}) are not exclusively modeled through $h_c(\cdot)$. Indeed, it is interesting to note, that the graphical decomposition provided by $\cal{C}$ leads to interactions which are different from the interactions the same decomposition would provide when used in a probabilistic graphical model, like for instance a Markov Random Field (MRF). In particular, a decomposition into unary terms only, as given in equation (\ref{eq:point_net}), does \emph{not} lead to independence between items, whereas an MRF with unary terms only is equivalent to a distribution over independent random variables. This is a consequence of the global mapping $f_\phi(\cdot)$, which is defined on the sum over all direct interactions. Higher-order interactions between several items not directly modeled through a non-linearity $h_c(\cdot)$ can eventually be learned by the model through the joint output space of all $h_c(\cdot)$, provided that the dimensionality $H$ of this space is high enough to incorporate all interactions. However, whereas the mapping $h_c(\cdot)$ provides a direct model of interactions between pairs of items, learning interactions between two items $(j,k)$, which are not directly captured through a clique $c$ and its corresponding $h_c(\cdot)$, requires learning a corresponding subspace in the common output space spanned by all $h_c(\cdot)$.

This leads to the question of how to define the trade-off between the complexity of the decomposition $\cal{C}$ and the output dimensionality $H$ of the mapping $h_c(\cdot)$, both of which will determine the complexity of the modeled interactions. Increasing the size of cliques in $\cal{C}$ will increase the input dimension (and therefore the capacity) of the mapping $h_c(\cdot)$ as well as the computational complexity of the sum operation.

\begin{figure}[t]
	\centering
	\includegraphics[width=9cm]{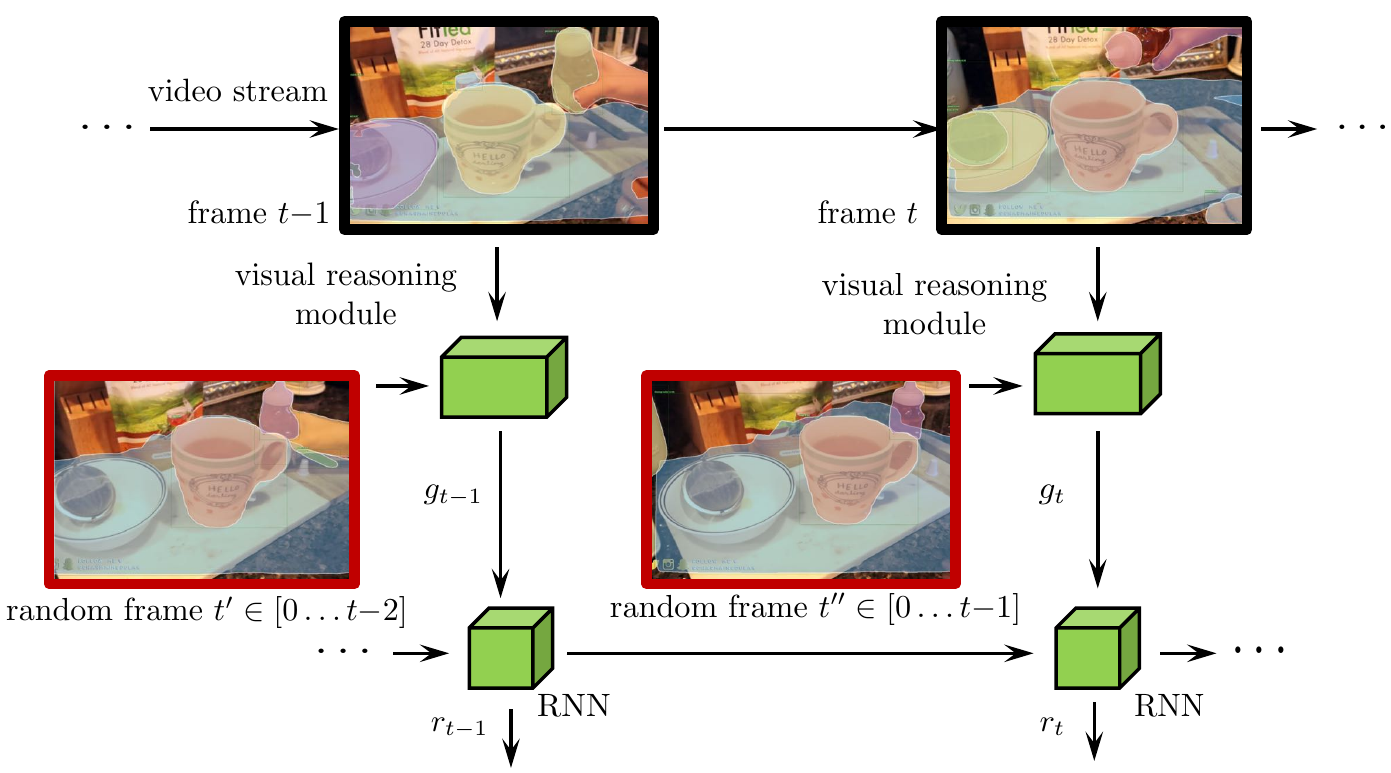}
	\caption{ORN in the object head operating on detected instances of objects.}
	\label{fig:orn}
\end{figure}

Inspired by relational networks~\cite{Santoro_2107_NIPS}, we chose to directly model inter-frame interactions between pairs of objects $(j,k)$ and leave modeling of higher-order interactions to the output space of the mappings $h_\theta$ and the global mapping $f_\phi$:
\begin{gather}
	\mg_t = \sum_{j,k} h_{\theta}(\mo^{j}_{t'},\mo^{k}_{t})
	\label{eq:g_t}
\end{gather}
In order to better directly model long-range interactions, we make the global mapping $f_\phi(\cdot,\cdot)$ recurrent, which leads to the following form:
\begin{gather}
	\mr_{t} =  f_\phi( \mg_t, \mr_{t-1} ) \label{eq:ovrn}
\end{gather}
where $\mr_t$ represents the recurrent \emph{object reasoning state} at time $t$ and $\mg_t$ is the global inter-frame interaction inferred at time $t$ such as described in Equation~\ref{eq:g_t}. In practice, this is implemented as a GRU, but for simplicity we omitted the gates in Equation (\ref{eq:ovrn}). The pairwise mappings $h_{\theta}(\cdot,\cdot)$ are implemented as an MLP. 
Figure \ref{fig:orn} provides a visual explanation of the object head's operating through time.

Our proposed ORN differs from \cite{Santoro_2107_NIPS} in three main points:

\noindent
\textbf{Objects have a semantic definition} --- we model relationships with respect to semantically meaningful entities (object instances) instead of feature map cells which do not have a semantically meaningful spatial extent. We will show in the experimental section that this is a key difference.

\noindent
\textbf{Objects are selected from different frames} --- we infer object pairwise relations only between objects present in two different sets. This is a key design choice which allows our model to reason about changes in object relationships over time. 

\noindent
\textbf{Long range reasoning} --- integration of the object relations over time is recurrent by using a RNN for $f_\phi(\cdot)$. Since reasoning from a full sequence cannot be done by inferring the relations between two frames, $f_\phi(\cdot)$ allows long range reasoning on sequences of variable length.

\subsection{Object instance features}
\label{sec:mask_predictor}
\noindent
The object features $\mO_{t} = \big\{ \mo^k_{t} \big\}^{K}_{k=1}$ for each frame $t$ used for the ORN module described above are computed and collected from local regions predicted by a mask predictor. Independently for each frame $\mX_t$ of the input data block, we predict object instances as binary masks $\mB_t^k$ and associated object class predictions $\mc_t^k$, a distribution over $C$ classes. We use Mask-RCNN~\cite{He_2017_ICCV}, which is able to detect objects in a frame using region proposal networks~\cite{Ren_2015_NIPS} and produces a high quality segmentation mask for each object instance. 

The objective is to collect features for each object instance, which jointly describe its appearance, the change in its appearance over time, and its shape, i.e. the shape of the binary mask. In theory, appearance could also be described by pooling the feature representation learned by the mask predictor (Mask R-CNN). However, in practice we choose to pool features from the dedicated \textit{object head} such as shown in Figure ~\ref{fig:block_diagram}, which also include motion through the spatio-temporal convolutions shared with the activity head:
\begin{gather}
\mathu_t^k = \text{ROI-Pooling} (\mU_t,\mB_t^k) 
\label{eq:feature_object}
\end{gather}
where $\mU_t$ is the feature map output by the \textit{object head}, $\mathu_t^k$ is a $D$-dimensional vector of appearance and appearance change of object $k$.


Shape information from the binary mask $\mB_t^k$ is extracted through the following mapping function: 
$
\mb_t^k = g_{\phi}(\mB_t^k),
$
where $g_{\phi}(\cdot)$ is a MLP.
Information about object $k$ in image $\mX_t$ is given by a concatenation of appearance, shape, and object class:
$
\mo_t^k = [ \ \mb_t^k \ \ \mathu_t^k \ \ \mc_t^k \ ]
$.

\subsection{Global Motion and Context}
\label{sec:context}
\noindent
Current approaches in video understanding focus on modeling the video from a high-level perspective.
By a stack of spatio-temporal convolution and pooling they focus on learning global scene context information.
Effective activity recognition requires integration of both of these sources: global information about the entire video content in addition to relational reasoning for making fine distinctions regarding object interactions and properties.



In our method, local low-level reasoning is provided through object head and the ORN module such as described above in Section \ref{sec:sorn}.
We complement this representation by high-level context information described by $\mV_{t}$ which are feature outputs from the activity head (orange block in Figure \ref{fig:block_diagram}).

We use spatial global average pooling over $\mV_t$ to output $T$ $D$-dimensional feature vectors denoted by $\mv_t$, where
$\mv_t$ corresponds to the context information of the video at timestep $t$.

We model the dynamics of the context information through time by employing a RNN $f_{\gamma}(\cdot)$ given by:
\begin{equation}
	\ms_t = f_{\gamma}(\mv_t, \ms_{t-1})
\end{equation}
where $\ms$ is the hidden state of $f_{\gamma}(\cdot)$ and gives cues about the evolution of the context though time.




\subsection{Recognition}
\noindent

\noindent
Given an input video sequence $\mX_{1:T}$, the two different streams corresponding to the activity head and the object head result in the two representations $\mh$ and $\mr$, respectively where $\mh = \sum_t $\mh$_t$ and $\mr = \sum_t $\mr$_t$. Each representation is the hidden state of the respective GRU, which were described in the preceding subsections. Recall that $\mh$ provides the global motion context while $\mr$ provides the object reasoning state output by the ORN module.
We perform independent linear classification for each representation:
\begin{gather}
	\my^1 = \mW\,\mh \label{eq:recognition_context} \\
	\my^2 = \mZ\,\mr \label{eq:recognition_reasoning}
\end{gather}
where $\my^1, \my^2$ correspond to the logits from the \textit{activity head} and the \textit{object head}, respectively, and $\mW$ and $\mZ$ are trainable weights (including biases). The final prediction is done by averaging logits $\my^1$ and $ \my^2$ followed by softmax activation.

\section{Network Architectures and feature dimensions}
\label{sec:architectures}

\noindent
The input RGB images $\mX_t$ are of size $\mathbf{R}^{3 \times W \times H}$ where $W$ and $H$ correspond to the width and height and are of size 224 each.
The object and activity heads (orange and green in Figure \ref{fig:block_diagram}) are a joint convolutional neural network with Resnet50 architecture pre-trained on ImageNet/ILSVRC \cite{Russakovsky2015}, with Conv1 and Conv5 blocks being inflated to 2.5D convolutions~\cite{Xie_arxiv_2017_a} (3D convolutions with a separable temporal dimension). This choice has been optimized on the validation set, as explained in Section \ref{sec:experiments} and shown in Table \ref{table:effect_arch_cnn}.

The last \textit{conv5} layers have been split into two different heads (activity head and object head).
The intermediate feature representations $\mU_t$ and $\mV_t$ are of dimensions $2048{\times}T{\times}14{\times}14$ and $2048{\times}T{\times}7{\times}7$, respectively.
We provide a higher spatial resolution for the feature maps $\mU_t$ of the object head to get more precise local descriptors.
This can be done by changing the stride of the initial \textit{conv5} layers from 2 to 1.
Temporal convolutions have been configured to keep the same time temporal dimension through the network.

Global spatial pooling of activity features results in a 2048 dimensional feature vector fed into a GRU with 512 dimensional hidden state $\ms_t$.
ROI-Pooling of object features results in 2048 dimensional feature vectors $\mathu_t^k$. 
The encoder of the binary mask is a MLP with one hidden layer of size 100 and outputs a mask embedding $\mb_t^k$ of dimension 100.
The number of object classes is 80, which leads in total to a 2229 dimensional object feature vector $\mo_t^k$.

The non-linearity $h_\theta(\cdot)$ is implemented as an MLP with 2 hidden layers each with 512 units and produces an 512 dimensional output space. $f_\phi(\cdot)$ is implemented as a GRU with a 256 dimension hidden state $\mr_t$.
We use ReLU as the activation function after each layer for each network.

\section{Training}
\label{sec:training}

\noindent
We train the model with a loss split into two terms:
\begin{equation}
\mathcal{L}_{tot} = 
\mathcal{L} \Big(\frac{\hat{\my}^1+\hat{\my}^2}{2},\my \Big)
 + \sum_{t}\sum_{k}\mathcal{L} (\hat{\mc}_t^k, \mc_t^k ).
\end{equation}
where $\mathcal{L}$ is the cross-entropy loss. 
The first term corresponds to supervised activity class losses comparing two different activity class predictions to the class ground truth: $\hat{\my}^1$ is the prediction of the activity head, whereas $\hat{\my}^2$ is the prediction of the object head, as given by Equations (\ref{eq:recognition_context}) and (\ref{eq:recognition_reasoning}), respectively.

The second term is a loss which pushes the features $\mU$ of the object towards representations of the semantic object classes. The goal is to obtain features related to, both, motion (through the layers shared with the activity head), as well as as object classes. As ground-truth object classes are not available, we define the loss as the cross-entropy between the class label $\mc_t^k$ predicted by the mask predictor and a dedicated linear class prediction $\hat{\mc}_t^k$ based on features $\mathu_t^k$, which, as we recall, are RoI-pooled from $\mU$:
\begin{equation}
	\mc_t^k = \mR~ \mathu_t^k
	\label{eq:object_class_mapping}
\end{equation}
where $\mR$ trainable parameters (biases integrated) learned end-to-end together with the other parameters of the model.

We found that first training the object head only and then the full network was performing better.
A ResNet50 network pretrained on ImageNet is modified by inflating some of its filters to 2.5 convolutions (3D convolutions with the time dimension separated), as described in Section \ref{sec:architectures}; then by fine-tuning.

We train the model using the Adam optimizer \cite{AdamOptimization2015} with an initial learning rate of $10^{-4}$ on 30 epochs and use early-stopping criterion on the validation set for hyper-parameter optimization. Training takes $\sim$50 minutes per epoch on 4 Titan XP GPUs with clips of 8 frames.

\section{Experimental results}
\label{sec:experiments}

We evaluated the method on three standard datasets, which represent difficult fine-grained activity recognition tasks: the Something-Something dataset, the VLOG dataset and the recently released EPIC Kitchens dataset.\smallskip

\myparagraph{\textbf{Something-Something (SS)} }%
is a recent video classification dataset with 108,000 example videos and 157 classes~\cite{Goyal_2017_ICCV}. It shows humans performing different actions with different objects, actions and objects being combined in different ways. 
Solving SS requires common sense reasoning and the state-of-the-art methods in activity recognition tend to fail, which makes this dataset challenging.\smallskip

\myparagraph{\textbf{VLOG}} is a multi-label binary classification of human-object interactions recently released with ~114,000 videos and 30 classes~\cite{Fouhey_2017_arxiv_b}.
Classes correspond to objects, and labels of a class are 1 if a person has touched a certain object during the video, otherwise they are 0. It has recently been shown, that state-of-the-art video based methods~\cite{Carreira_2017_CVPR} are outperformed on VLOG by image based methods like ResNet-50~\cite{He2015}, although these video methods outperform image based ResNet-50 on large-scale video datasets like the Kinetics dataset~\cite{Carreira_2017_CVPR}. This suggests a gap between traditional datasets like Kinetics and the fine-grained dataset VLOG, making it particularly difficult.\smallskip

\myparagraph{\textbf{EPIC Kitchens (EPIC)} }%
is an egocentric video dataset recently released containing 55 hours recording of daily activities~\cite{Damen2018EPICKITCHENS}.
This is the largest in first-person vision and the activities performed are non-scripted, which makes the dataset very challenging and close to real world data.
The dataset is densely annotated and several tasks exist such as object detection, action recognition and action prediction.
We focus on action recognition with 39'594 action segments in total and 125 actions classes (i.e verbs).
Since the test set is not available yet we conducted our experiments on the training set (28'561 videos).
We use the videos recorded by person 01 to person 25 for training (22'675 videos) and define the validation set as the remaining videos (5'886 videos).\smallskip

For all datasets we rescale the input video resolution to $256{\times}256$. While training, we crop space-time blocks of $224{\times}224$ spatial resolution and $L$ frames, with $L{=}8$ for the SS dataset and $L{=}4$ for VLOG and EPIC.
We do not perform any other data augmentation.
While training we extract $L$ frames from the entire video by splitting the video into $L$ sub-sequences and randomly sampling one frame per sub-sequence.
The output sequence of size $L$ is called a \textit{clip}.
A clip aims to represent the full video with less frames.
For testing we aggregate results of 10 clips.
We use \textit{lintel}~\cite{lintel} for decoding video on the fly.

The ablation study is done by using the train set as training data and we report the result on the validation set.
We compare against other state-of-the-art approaches on the test set. For the ablation studies, we slightly decreased the computational complexity of the model: the base network (including activity and object heads) is a ResNet-18 instead of ResNet-50, a single clip of 4 frames is extracted from a video at test time.

\myparagraph{Comparison with other approaches. }%
Table \ref{table:sota_vlog} shows the performance of the proposed approach on the VLOG dataset. We outperform the state of the art on this challenging dataset by a margin of $\approx$4.2 points (44.7\% accuracy against 40.5\% by~\cite{He2015}). As mentioned above, traditional video approaches tend to fail on this challenging fine-grained dataset, providing inferior results. Table \ref{table:sota_something} shows performance on SS where we outperform the state of the art given by very recent methods (+2.3 points).
On EPIC we re-implement standard baselines and report results on the validation set (Table \ref{table:sota_epic}) since the test set is not available.
Our full method reports an accuracy of 40.89 and outperforms baselines by a large margin ($\approx$+6.4 and $\approx$+7.9 points respectively for against CNN-2D and I3D based on a ResNet-18).\smallskip

\begin{table*} [t]
\caption{{ Results on Hand/Semantic Object Interaction Classification (Average precision in \% on the test set) on VLOG dataset. R50 and I3D implemented by~\cite{Fouhey_2017_arxiv_b}.}}
 \resizebox{\textwidth}{!}{
\begin{tabular}{l@{  }c@{  }c@{  }c@{  }c@{  }c@{  }c@{  }c@{  }c@{  }c@{  }c@{  }c@{  }c@{  }c@{  }c@{  }c@{  }c@{  }c@{  }c@{  }c@{  }c@{  }c@{  }c@{  }c@{  }c@{  }c@{  }c@{  }c@{  }c@{  }c@{  }c@{  }c}
\toprule
  & \small \rotatebox{90}{mAP}  & \small \rotatebox{90}{bag} 
 & \small \rotatebox{90}{bed} 
 & \small \rotatebox{90}{bedding} 
 & \small \rotatebox{90}{book/papers} 
 & \small \rotatebox{90}{bottle/tube} 
 & \small \rotatebox{90}{bowl} 
 & \small \rotatebox{90}{box} 
 & \small \rotatebox{90}{brush} 
 & \small \rotatebox{90}{cabinet} 
 & \small \rotatebox{90}{cell-phone} 
 & \small \rotatebox{90}{clothing} 
 & \small \rotatebox{90}{cup} 
 & \small \rotatebox{90}{door} 
 & \small \rotatebox{90}{drawers} 
 & \small \rotatebox{90}{food} 
 & \small \rotatebox{90}{fork} 
 & \small \rotatebox{90}{knife} 
 & \small \rotatebox{90}{laptop} 
 & \small \rotatebox{90}{microwave} 
 & \small \rotatebox{90}{oven} 
 & \small \rotatebox{90}{pen/pencil} 
 & \small \rotatebox{90}{pillow} 
 & \small \rotatebox{90}{plate} 
 & \small \rotatebox{90}{refrigerator} 
 & \small \rotatebox{90}{sink} 
 & \small \rotatebox{90}{spoon} 
 & \small \rotatebox{90}{stuffed animal} 
 & \small \rotatebox{90}{table} 
 & \small \rotatebox{90}{toothbrush} 
 & \small \rotatebox{90}{towel} 
\\ 
\midrule\small R50~\cite{He2015}  & \scriptsize 40.5  & \scriptsize 29.7  & \scriptsize 68.9  & \scriptsize 65.8  & \scriptsize  64.5  & \scriptsize 58.2  & \scriptsize  33.1  & \scriptsize  22.1  & \scriptsize 19.0  & \scriptsize  \bf 23.9  & \scriptsize  54.0  & \scriptsize 45.5  & \scriptsize 28.6  & \scriptsize 49.2  & \scriptsize \bf 28.7  & \scriptsize 49.6  & \scriptsize 19.4  & \scriptsize 37.5  & \scriptsize  62.9  & \scriptsize  48.8  & \scriptsize  23.0  & \scriptsize 36.9  & \scriptsize 39.2  & \scriptsize 12.5  & \scriptsize \bf 55.9  & \scriptsize 58.8  & \scriptsize 31.1  & \scriptsize \bf 57.4  & \scriptsize  26.8  & \scriptsize 39.6  & \scriptsize 22.9 \\
\small I3D~\cite{Carreira_2017_CVPR} & \scriptsize 39.7  & \scriptsize 24.9  & \scriptsize  71.7  & \scriptsize \bf 71.4  & \scriptsize 62.5  & \scriptsize 57.1  & \scriptsize 27.1  & \scriptsize 19.2  & \scriptsize \bf 33.9  & \scriptsize 20.7  & \scriptsize 50.6  & \scriptsize 45.8  & \scriptsize 24.7  & \scriptsize  54.7  & \scriptsize 19.1  & \scriptsize \bf 50.8  & \scriptsize 19.3  & \scriptsize \bf 41.9  & \scriptsize 54.0  & \scriptsize 27.5  & \scriptsize 21.4  & \scriptsize  37.4  & \scriptsize 42.9  & \scriptsize 12.6  & \scriptsize 42.5  & \scriptsize  60.4  & \scriptsize  33.9  & \scriptsize 46.0  & \scriptsize 23.5  & \scriptsize  59.6  & \scriptsize  34.7 \\
\small \bf Ours  & \scriptsize \bf 44.7  & \scriptsize  \bf 30.2  & \scriptsize \bf 72.3  & \scriptsize 70.7  & \scriptsize \bf 64.9  & \scriptsize \bf 59.8  & \scriptsize \bf 38.2  & \scriptsize  \bf 24.6  & \scriptsize 26.3 & \scriptsize 22.4  & \scriptsize \bf 64.5  & \scriptsize \bf 47.2  & \scriptsize \bf 35.4  & \scriptsize \bf 57.9  & \scriptsize  25.2  & \scriptsize 48.5  & \scriptsize \bf 24.5 & \scriptsize 40.2  & \scriptsize \bf 72.0  & \scriptsize  \bf 54.1  & \scriptsize \bf 26.5  & \scriptsize \bf 39.9  & \scriptsize \bf 48.6  & \scriptsize \bf 15.2  & \scriptsize 53.5 & \scriptsize \bf 60.7  & \scriptsize \bf 36.8  & \scriptsize 52.8  & \scriptsize \bf 27.9 & \scriptsize \bf 64.0  & \scriptsize \bf 37.6 \\
\bottomrule 
\end{tabular}
}

\label{table:sota_vlog}
\end{table*}

\myparagraph{Effect of object-level reasoning. }%
Table~\ref{table:ablation_heads} shows the importance of reasoning on the performance of the method. 
The baseline corresponds to the performance obtained by the activity head trained alone (inflated ResNet, in the ResNet-18 version for this table). No object level reasoning is present in this baseline. The proposed approach (third line) including an object head and the ORN module gains 0.8, 2.5 and 2.4 points compared to our baseline respectively on SS,  on EPIC and on VLOG. This indicates that the reasoning module is able to extract complementary features compared to the activity head.

Using \textit{semantically defined objects} proved to be important and led to a gain of 2 points on EPIC and 2.3 points on VLOG for the full model (6/12.7 points using the object head only) compared to an extension of Santoro \etal~\cite{Santoro_2107_NIPS} operating on pixel level.
This indicates importance of object level reasoning.
The gain on SS is smaller (0.7 point with the full model and 7.8 points with the object head only) and can be explained by the difference in spatial resolution of the videos. Object detections and predictions of the binary masks are done using the initial video resolution. The mean video resolution for VLOG is $660{\times}1183$ and for EPIC is $640{\times}480$ against $100{\times}157$ for SS.
Mask-RCNN has been trained on images of resolution $800{\times}800$ and thus performs best on higher resolutions.
The quality of the object detector is important for leveraging object level understanding then for the rest of the ablation study we focus on EPIC and VLOG datasets.





The function $f_\phi$ in Equation (\ref{eq:ovrn}) is an important design choice in our model. In our proposed model, $f_\phi$ is recurrent over time to ensure that the ORN module captures long range reasoning over time, as shown in Equation (\ref{eq:ovrn}). Removing the recurrence in this equation leads to an MLP instead of a (gated) RNN, as evaluated in row 4 of Table~\ref{table:ablation_heads}. Performance decreases by 1.1 point on VLOG and 1.4 points on EPIC. The larger gap for EPIC compared to VLOG and can arguably be explained by the fact that in SS actions cover the whole video, while 
solving VLOG requires detecting the right moment when the human-object interaction occurs and thus long range reasoning plays a less important role.

Visual features extracted from object regions are the most discriminative, however object shapes and labels 
also provide complementary information.
Finally, the last part of Table~\ref{table:ablation_heads} evaluates the effect of the cliques size for modeling the interactions between objects and show that pairwise cliques outperform cliques of size 1 and 3.
We would like to recall, that even with unary cliques only, interactions between objects are still modeled. However, the model needs to find subspaces in the hidden representations associated to each interaction.

\begin{table*}[t]
   \begin{center}
		\caption{Ablation study with ResNet-18 backbone. Results in \%: Top-1 accuracy for EPIC and SS datasets, and mAP for VLOG dataset.}
		\scalebox{0.9}{
\begin{tabular}{c|c|cc|cc|cc}
	\arrayrulecolor{tablecolor} \toprule
	\multirow{2}{*}{Method} & \multirow{2}{*}{Object type}	 &  \multicolumn{2}{c|}{\,EPIC} & \multicolumn{2}{c|}{VLOG} &  \multicolumn{2}{c}{\,SS}			\\
					   		& 			 					&  \,obj. head 			& 2 heads &  \,obj. head 			& 2 heads  			&   \,obj. head 	 	& 2 heads 	\\
	\arrayrulecolor{tablecolor} \toprule
	\textit{Baseline}		   		& -  				 		   	&		-	   			&	\textit{38.33}    			&		-				&	\textit{35.03}	& -				&	\textit{31.31}	\\
	ORN				   		& pixel 						&		23.71			&	38.83    		 	&		14.40			&	35.18	&  2.51 &	31.43 \\
	\textbf{ORN}				   		& \textbf{COCO}  						&		\textbf{29.94}	&	\textbf{40.89}    	&		\textbf{27.14}	&	\textbf{37.49} &  10.26 &	\textbf{32.12}\\ %
	ORN-mlp				   		& COCO 						&		28.15	&	39.41	&	25.40		&	36.35 \vspace{-0.5mm} &-&- \\ %
	\midrule\small
	ORN				   		& \,COCO-visual\,		  		  	&		28.45			&	38.92    			&		22.92			&	35.49	&-&- 	\\
	ORN				   		& COCO-shape  			   		&		21.92			&	37.16    			&		7.18			&	35.39	&-&- 	\\
	ORN				   		& COCO-class			  		&		21.96			&	37.75    			&		13.40			&	35.94		&-&- \\ %
	\midrule
	ORN		   		& COCO-intra  						&		29.25			&	38.10    			&		26.78			&	36.28		&-&- \\ 
	ORN	clique{-}1		   		& COCO  						&		28.25			&	40.18    			&		26.48			&	36.71		&-&- \\
	ORN	clique{-}3\,		   		& COCO  						&		22.61			&	37.67				&		27.05			&	36.04		&-&- \\
    \arrayrulecolor{tablecolor} \bottomrule
\end{tabular}
}

		\label{table:ablation_heads}
    \end{center}
\end{table*}

\begin{table*}[t]
	\begin{center}
		\begin{tabular}{ccc}
	\begin{minipage}{.44\linewidth}
			\caption{Experimental results on the Something-Something dataset (classification accuracy in \% on the test set).}
		\centering
\scalebox{0.9}{
		\begin{tabular}{c|c}
			\arrayrulecolor{tablecolor} \toprule
			Methods 													& 			\;\;\;Top1\;\;\; 				\\
			 \midrule
			C3D + Avg~\cite{Goyal_2017_ICCV}							& 			21.50		  		\\
			I3D~\cite{Goyal_2017_ICCV}									& 			27.63		  		\\	
			MultiScale TRN~\cite{Zhou_2017}\;\;							& 			33.60		  		\\
			\midrule\small
			\bf Ours													& 			 \textbf{35.97}			  	\\	
            \arrayrulecolor{tablecolor} \bottomrule
		\end{tabular}
}
		\label{table:sota_something}
	\end{minipage}\;\;\;\;\;\;\;\;\;\;\;\;\;\;%
	&&
	\begin{minipage}{.44\linewidth}
		\caption{Experimental results on the EPIC Kitchens dataset (accuracy in \% on the validation set -- methods with~$^*$ have been re-implemented).}
		\centering
\scalebox{0.9}{
		\begin{tabular}{c|c}
			\arrayrulecolor{tablecolor} \toprule
			Methods 													& 			\;\;\;Top1\;\;\; 				\\
			\midrule
			R18 ~\cite{He2015}$^*$						& 			32.05		  		\\
			I3D-18~\cite{Carreira_2017_CVPR}$^*\;$									& 			34.20		  		\\	
			\midrule\small
			\bf Ours													& 			 \textbf{40.89}			  	\\	
            \arrayrulecolor{tablecolor} \bottomrule
		\end{tabular}
}
   		\label{table:sota_epic}
	\end{minipage}\\
	\end{tabular}
	\end{center}
\end{table*}

\myparagraph{CNN architecture and kernel inflations. }%
The convolutional architecture of the model was optimized over the validation set of the SS dataset, as shown in Table~\ref{table:effect_arch_cnn}. The architecture itself (in terms of numbers of layers, filters etc.) is determined by pre-training on image classification. We optimized the choice of filter inflations from 2D to 2.5D or 3D for several convolutional blocks. This has been optimized for the single head model and using a ResNet-18 variant to speed up computation.
Adding temporal convolutions increases performance up to 100\% w.r.t. to pure 2D baselines.
This indicates, without surprise, that motion is a strong cue. Inflating kernels to 2.5D on the input side and on the output side provided best performances, suggesting that temporal integration is required at a very low level (motion estimation) as well as on a very high level, close to reasoning. Our study also corroborates recent research in activity recognition, indicating that 2.5D kernels provide a good trade-off between high-capacity and learnable numbers of parameters.
Finally temporal integration via RNN outperforms global average pooling over space and time.
The choice of a (gated) RNN for temporal integration of the activity head features proved important (see Table~\ref{table:effect_arch_cnn}) compared to global average pooling (GAP) over space and time.

\begin{table*}[t]
	\begin{center}
        \caption{Effect of the CNN architecture (choice of kernel inflations) on a single head ResNet-18 network. Accuracy in \% on the validation set of Something-Something is shown. 2.5D kernels are separable kernels: 2D followed by a 1D temporal.}
%
\resizebox{0.9\textwidth}{!}{
		\begin{tabular}{ccc|ccc|ccc|ccc|ccc|cc|cc}
			\arrayrulecolor{tablecolor} \toprule
			\multicolumn{3}{c}{Conv1} 	& \multicolumn{3}{c}{Conv2} 	& \multicolumn{3}{c}{Conv3} 	& \multicolumn{3}{c}{Conv4} 	& \multicolumn{3}{c}{Conv5} & \multicolumn{2}{c}{Aggreg}	& \multirow{2}{*}{\;\;\;SS\;\;\;}\\
			\small \;2D & 3D & 2.5D\; & \;2D & 3D & 2.5D\; & \;2D & 3D & 2.5D\; & \;2D & 3D & 2.5D\; & \;2D & 3D & 2.5D\; &\;GAP & RNN\; \\
			\arrayrulecolor{tablecolor} \toprule
			\checkmark & - & - & \checkmark & - & - & \checkmark & - & - & \checkmark & - & - & \checkmark & - & - & \checkmark & - & 15.73		\\
			\checkmark & - & - & \checkmark & - & - & \checkmark & - & - & \checkmark & - & - & \checkmark & - & - & - & \checkmark & 15.88	\\
			- & \checkmark & - & - & \checkmark & - & - & \checkmark & - & - & \checkmark & - & - & \checkmark  & - & \checkmark & - & 31.42	\\
			- & - & \checkmark & - & - & \checkmark & - & - & \checkmark & - & - & \checkmark & - & -  & \checkmark & \checkmark & - & 27.58 	\\
		\midrule\small
			\checkmark & - & - & \checkmark & - & - & \checkmark & - & - & \checkmark & - & - & - & \checkmark & - & \checkmark & - & 31.28 	\\
			\checkmark & - & - & \checkmark & - & - & \checkmark & - & - & - & \checkmark & - & - & \checkmark & - & \checkmark & - & 32.06 	\\
			\checkmark & - & - & \checkmark & - & - & - & \checkmark & - & - & \checkmark & - & - & \checkmark & - & \checkmark & - & 32.25 \\
			\checkmark & - & - & \checkmark & - & - & \checkmark & - & - & \checkmark & - & - & - & - & \checkmark & \checkmark & - & 31.31 	\\
			\checkmark & - & - & \checkmark & - & - & \checkmark & - & - & - & - & \checkmark  & - & - & \checkmark & \checkmark & - & 32.79 \\
			\checkmark & - & - & \checkmark & - & - & - & - & \checkmark & - & - & \checkmark  & - & - & \checkmark & \checkmark & - & \textbf{33.77} 	\\
			\midrule\small

			- & \checkmark & - & \checkmark & - & - & \checkmark & - & - & \checkmark & - & - & \checkmark & - & - & \checkmark & - & 28.71 	\\
			- & \checkmark & - & - & \checkmark  & - & \checkmark & - & - & \checkmark & - & - & \checkmark & - & - & \checkmark & - &  31.42 \\

			- & - & \checkmark & \checkmark & - & - & \checkmark & - & - & \checkmark & - & - & \checkmark & - & - & \checkmark & - & 20.05 	\\
			- & - & \checkmark & - & - & \checkmark & \checkmark & - & - & \checkmark & - & - & \checkmark & - & - & \checkmark & - & 22.52 	\\
			\arrayrulecolor{tablecolor} \bottomrule
		\end{tabular}
}

   	\label{table:effect_arch_cnn}
	\end{center}
\end{table*}

\begin{figure*}[t!] \centering
\includegraphics[width=\linewidth]{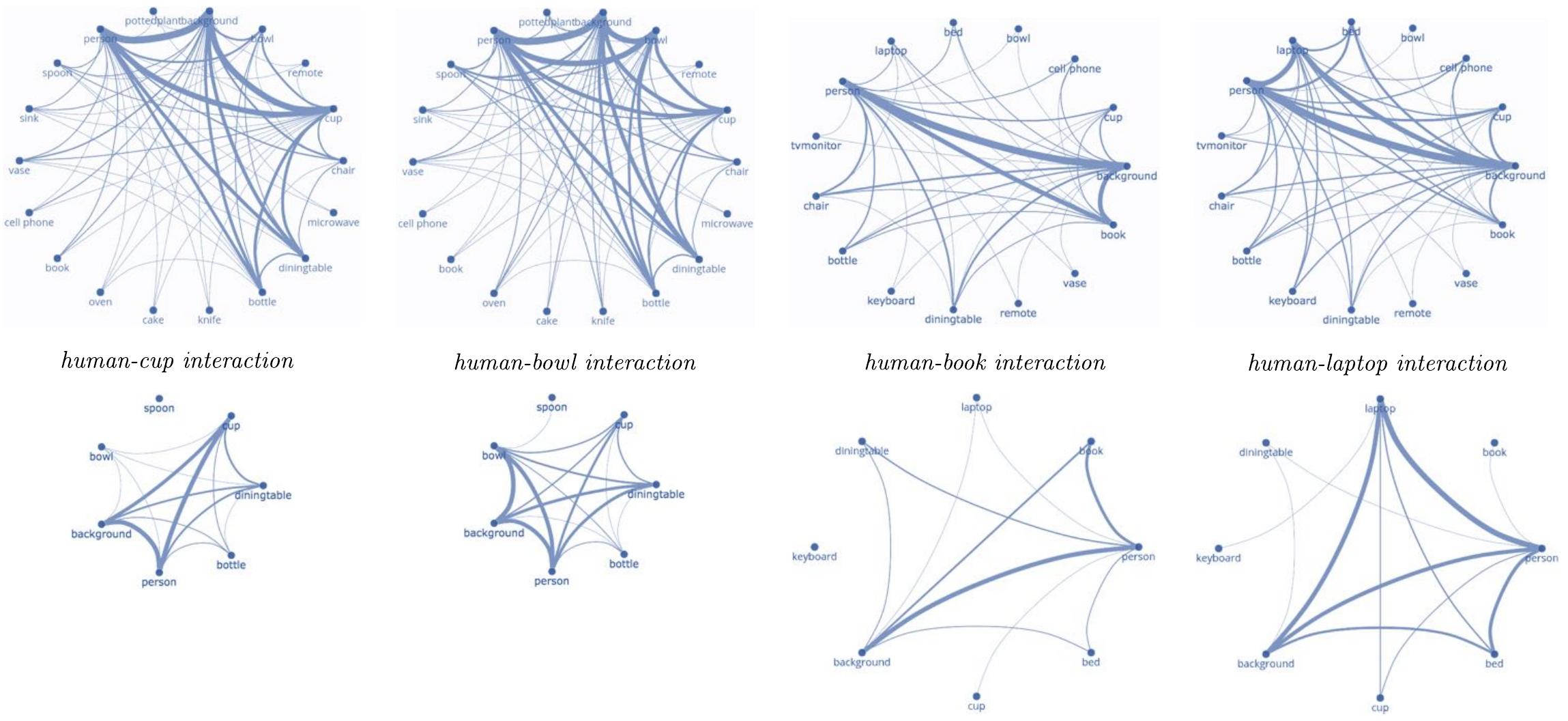}
\caption{\label{fig:interactions} Example of object pairwise interactions learned by our model on VLOG for four different classes. 
Objects co-occurrences are at the top and learned pairwise objects interactions are at the bottom. Line thickness indicates learned importance of a given relation. Interactions have been normalized by the object co-occurrences.} 
\end{figure*}


\myparagraph{Visualizing the learned object interactions. }%
Figure \ref{fig:interactions} shows visualizations of the pairwise object relationships the model learned from data, in particular from the VLOG dataset. Each graph is computed for a given activity class, and strong arcs between two nodes in the graph indicate strong relationships between the object classes, i.e. the model detects a high correlation between these relationships and the activity. The graphs were obtained by thresholding the summed activations of each pairwise relationship $(j,k)$ in equation (\ref{eq:g_t}). Each pair $(j,k)$ can be assigned a pair of object classes $\mc_t^j$ and $\mc_t^k$ through the predictions of the object instance mask predictor. Integrating over all samples of the dataset for a given class leads to the visualizations in Figure \ref{fig:interactions}.
We can see that the object interactions are highly relevant to the detected activities: the \emph{person-touches-bed} activity is correlated to interactions between relevant object classes \emph{person} and \emph{bed}. Similarly, activities \emph{human-bowl interaction} and \emph{human-cup interaction} show interactions with the respective objects \emph{bowl} and \emph{cup}. Moreover, other recovered relationships are highly correlated to the scene (for example, \emph{dining-table} and \emph{bowl} for activity \emph{human-bowl interaction}).

\begin{figure*}[t!] \centering
\includegraphics[width=6cm]{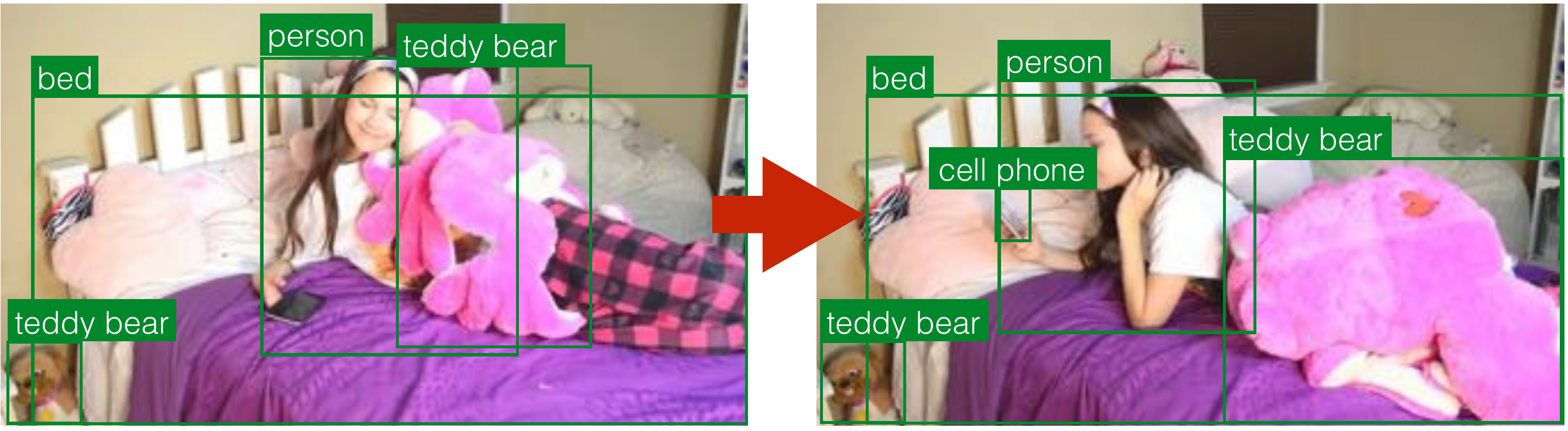}\hfill%
\includegraphics[width=6cm]{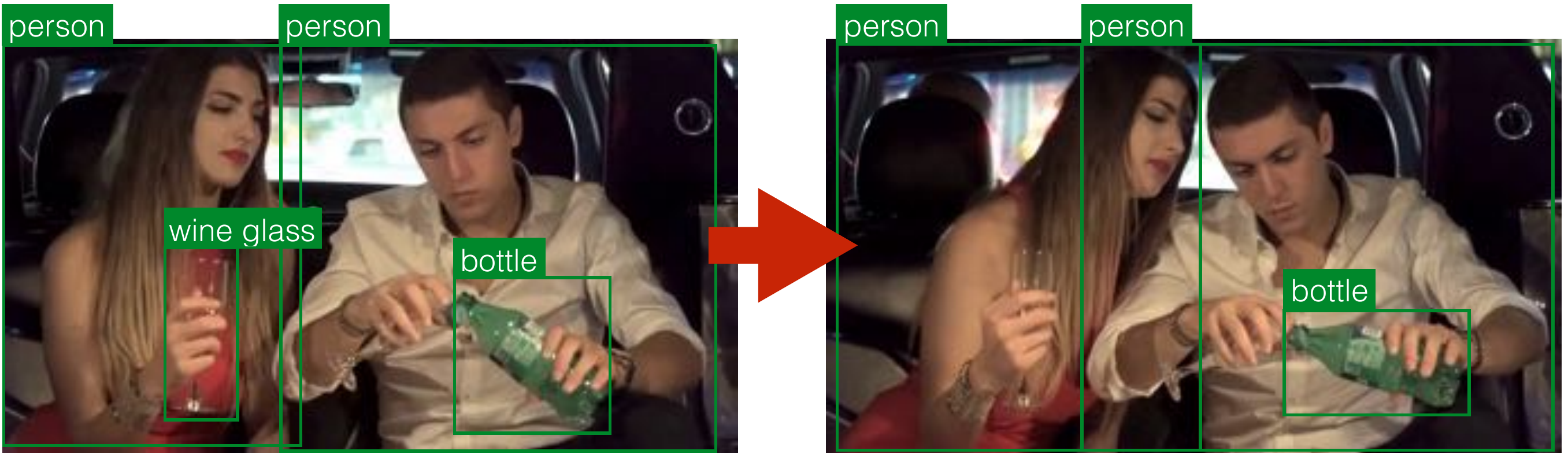}
\label{fig:failures}
\caption{\label{fig:failure_cases} \textbf{Examples of failure cases} -- a) small sized objects (on the left).
Our model detects a \textit{cell phone} and a \textit{person} but fails to detect \textit{hand-cell-phone contact}; 
b) confusion between semantically similar objects (on the right). The model falsly predicts \textit{hand-cup contact}  instead of \textit{hand-glass-contact} even though the \textit{wine glass} is detected.} 
\end{figure*}

Finally, Figure~\ref{fig:failure_cases} shows some failure cases, which are either due to errors done by the object mask prediction (Mask R-CNN) or by the ORN itself.

\section{Conclusion}
\label{sec:conclusion}
\noindent
We presented a method for activity recognition in videos which leverages 
object instance detections for visual reasoning on object interactions over time.
The choice of reasoning over semantically well-defined objects is key to our approach
and outperforms state of the art methods which reason on grid-levels,
such as cells of 
convolutional feature maps.
Temporal dependencies and causal relationships are dealt with by integrating relationships between different time instants.
We evaluated the method on three difficult datasets, on which standard approaches do not perform well, and report state-of-the-art results. 
\\

\noindent
\textbf{Acknowledgements.} This work was funded by grant Deepvision (ANR-15-CE23-0029, STPGP-479356-15), a joint French/Canadian call by ANR \& NSERC.

\bibliographystyle{splncs04}
\bibliography{refs}

\end{document}